\documentclass{article}

\usepackage{arxiv}

\usepackage[utf8]{inputenc} 
\usepackage[T1]{fontenc}    
\usepackage{hyperref}       
\usepackage{url}            
\usepackage{booktabs}       
\usepackage{amsfonts}       
\usepackage{amsmath}
\usepackage{nicefrac}       
\usepackage{microtype}      
\usepackage{lipsum}
\usepackage{threeparttable}
\usepackage{graphicx}
\graphicspath{ {./images/} }
\usepackage[style=nejm, 
citestyle=numeric-comp,
sorting=none]{biblatex}
\title{Machine Learning Approach for Cancer Entities Association and Classification}

\author{
 G.Jeyakodi \\
  Centre for Bioinformatics\\
  Pondicherry University\\
  Puducherry, India 605014 \\
  \texttt{rjeyakodi02@gmail.com} \\
   \And
 Arkadeep Pal \\
  Centre for Bioinformatics\\
  Pondicherry University\\
  Puducherry, India 605014 \\
  \texttt{palarkadeep97@gmail.com} \\
  \And
 Debapratim Gupta \\
  Department of Computer Science\\
  Pondicherry University\\
  Puducherry, India 605014 \\
  \texttt{debexam1234@gmail.com} \\
  \And
  K.Sarukeswari \\
   Department of Computer Science\\
   Pondicherry University\\
   Puducherry, India 605014 \\
  \texttt{sarukeswari@gmail.com} \\
  \And
  V.Amouda \\
   Centre for Bioinformatics\\
   Pondicherry University\\
   Puducherry, India 605014 \\
  \texttt{amouda@yahoo.com} \\
}

\begin{document}
\maketitle
\begin{abstract}
According to the World Health Organization (WHO), cancer is the second leading cause of
death globally. Scientific research on different types of cancers grows at an ever-increasing rate,
publishing large volumes of research articles every year. The insight information and the knowledge of the drug, diagnostics, risk, symptoms, treatments, etc., related to genes are significant
factors that help explore and advance the cancer research progression. Manual screening of such
a large volume of articles is very laborious and time-consuming to formulate any hypothesis.
The study uses the two most non-trivial NLP, Natural Language Processing functions, Entity
Recognition, and text classification to discover knowledge from biomedical literature. Named
Entity Recognition (NER) recognizes and extracts the predefined entities related to cancer from
unstructured text with the support of a user-friendly interface and built-in dictionaries. Text
classification helps to explore the insights into the text and simplifies data categorization, query-
ing, and article screening. Machine learning classifiers are also used to build the classification
model and Structured Query Languages (SQL) is used to identify the hidden relations that may
lead to significant predictions.
\end{abstract}


\section{Introduction}
As numerous biomedical research articles are published regularly, adding knowledge to the
accumulated literature on different diseases, such as cancer, neurodegenerative diseases, and
hereditary diseases. One of the leading causes of global mortality disease is cancer due to various
reasons such as lifestyle habits, radiation exposure, viral infections, and tobacco consumption \cite{mathur-sathishkumar2020} \cite{Sung2021-xp}
. These reasons ultimately make some genetic change in a
cell of tissue which causes it to become cancerous. Due to the top priority given to cancer research
compared to other human diseases, enormous articles were published \cite{Friedman2015-lf} \cite{Shipman2015-vi} in a short period \cite{talukder2022machine}. It can serve as a relevant source
for cancer knowledge discovery in different fields of diagnostics, application of drugs, genetic
association, prevention, and treatment. An automate downloading of articles and extraction of
related entities will advance the progression of the research faster.
Natural Language Processing (NLP) helps in communicating computers with humans in their
language and converts the unstructured data into structured data to improve the accuracy of text
mining. NLP function guides to understanding the human query language to discover knowledge
from literature without much manual effort \cite{Nadkarni2011}.
Named Entity Recognition (NER) and text classification is used mainly for text mining \cite{Khurana2023-cf}. Text mining identifies the meaningful patterns and new insights by transforming the
unstructured text into a structure. NER identifies the key features: genes, and tissue names crucial
to draw associations and inferences on a dataset under study.
The extracted named entities would not carry much meaning without a known label for a text \cite{joulin2016bag}. NER is one of the fundamental functions that require text classification to
uncover many relations that may be novel and have great importance in biological research \cite{eklund-usman2020}. Text classification is the task of assigning predefined categories to text
documents \cite{Aggarwal2012}. The main advantage of text classification is the automatic
prediction of a label for a text, a backbone of text mining.
The first step of the mining collects text articles as there are no pre-made datasets available for
dataset preparation. PubMed provides various E-utilities for biomedical text collection, one such
software, Entrez Direct allows users to download articles directly using different commands in
different formats available at National Center for Biotechnology Information - NCBI (nih.gov).
As the software is UNIX-based, it requires extensive knowledge to use various commands to
retrieve the articles and does not afford full-text articles with PubMed ID. The existing tools do
not provide the facility of extracting user-preferred entities from text documents. It is
advantageous to have user-level freedom to draw inter-entity associations such as gene-drug or
disease-gene associations \cite{gopal-jeyakodi-prakash2021}. It allows the user to
search for the associations related to a more specific entity or entity type.
Hence NLP and text mining are primary steps in extending research in any disease, more
mainstreaming cancer research. It also suggested an efficient, simple, comparatively small text
mining approach in formulating novel hypotheses from the associations mined from literature. To
overcome the above-mentioned drawbacks and due to the necessity of an automated tool, the
current study focuses on two tasks: 1. User interface development to extract the cancer entities
from the full-text articles. 2. Classification model developed to classify different types of cancers
using machine learning algorithms.
Many types of cancer entities like genes, proteins, drugs, and diagnostic methods are extracted
using NER, such types of named entities are equally important to other studies like point mutations,
chromosomal aberrations, symptoms, carcinogens, etc. A flexible dictionary-based approach to
extracting cancer entities \cite{Song2018-lb} to mine by preparing a dictionary of cancer entities
extracted from a text corpus.
The named entity extraction by dictionary-based approach does not need the knowledge of
complex NLP architecture and has good accuracy \cite{QUIMBAYA201655}. There have been many
advances that make the task of text mining easy and more accurate. The prominent advances are
the development of transformers \cite{vaswani-kaiser2017} using modern text mining architecture such
as BERT, a Bidirectional Encoder Representations from Transformers \cite{devlin2019bert}. The
text mining platform, WEKA (Waikato Environment for Knowledge Analysis) is helpful for data
preparation, classification, regression, clustering, association rules mining, and visualization \cite{Holmes1994-xl}. The toolkit SparkText \cite{Ye2016-cu} allows text mining
and text classification on big data architecture \cite{demchenko-yuri2014} for
biomedical articles. The extracted entities play a significant role in finding the hidden patterns, the
relationship between them helps in disease diagnosis, drug development, etc.\cite{Song2018-lb}
Associations describe a link between multiple items in a dataset that may be observable in the
hidden patterns or relationships that occur together \cite{ceglar-aaron2006}. Predicting
associations by filtering the observed patterns using a complex query and the hidden associations
using data mining algorithms such as Apriori or Eclat \cite{zaki2000}. These are beneficial in
extending and understanding cancer biology and discovering potential relationships among them.
It is not only restricted to cancer studies but also benefits all fields of biomedical studies.
This research developed an efficient pipeline using a machine learning algorithm to identify the
relationship among the named entities to predict the hidden patterns and a cancer classification
model.

\section{Materials and Methods}
\begin{figure}
    \centering
    \includegraphics{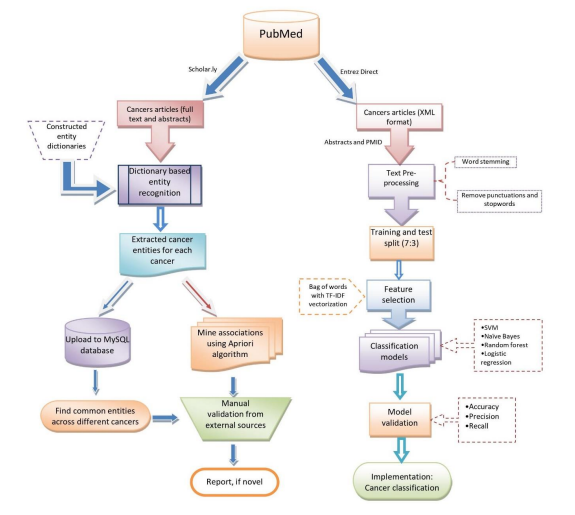}
  \caption{The basic workflow of the methodology}
  \label{fig:fig1}
\end{figure}

Figure \ref{fig:fig1} depicts the detailed workflow of the study by considering abstracts and full-text articles
for an automatic Named Entity Extraction (NER), and abstracts alone for cancer classification for
better accuracy \cite{Ye2016-cu}.

\subsection*{Data Collection}
Download the required articles for the mining of entities and classification. An automated tool,
Scholar.ly at https://github.com/Jeyakodi-Hub/Scholarly (a free open source in GitHub developed
by the same research group) is used to download the articles for oral cancer, cervical cancer, lung
cancer, gastric cancer, esophageal cancer, colorectal cancer, skin cancer, prostate cancer, breast
cancer, and pancreatic cancer. Total of 700 articles, out of which 70 articles for each cancer type
considered for NER to extract the entities. For the next task, text classification 46,000 abstracts
were downloaded using the query listed in Table \ref{tab:1} via Entrez Direct E-utility software from the
PubMed database.
\begin{table}[p]
    \caption{The PubMed queries made for searching articles and downloading via Entrez Direct}    
    
    \begin{tabular}{p{0.35\linewidth} | p{0.6\linewidth}}
            \hline
            Cancer type & PubMed queries \\  
            \hline
            oral cancer & (((oral cancer[MeSH Major Topic]) AND oral cancer[Title]) AND (‘2000/01/01’[Date - MeSH] :
‘2020/12/01’[Date - MeSH])) AND (English[Language])
 \\ \hline
            cervical cancer & (((cervical cancer[MeSH Major Topic]) AND cervical cancer[Title]) AND (‘2000/01/01’[Date -
MeSH] : ‘2020/12/01’[Date - MeSH])) AND (English[Language]) \\
            \hline
            lung cancer & (((Lung cancer[MeSH Major Topic]) AND Lung cancer[Title]) AND (‘2000/01/01’[Date - MeSH]
: ‘2020/12/01’[Date - MeSH])) AND (English[Language]) \\
            \hline
            gastric cancer & (((gastric cancer[MeSH Major Topic]) AND gastric cancer[Title]) AND (‘2000/01/01’[Date -
MeSH] : ‘2020/12/01’[Date - MeSH])) AND (English[Language]) \\
            \hline
            esophageal
cancer & (((Esophageal cancer[MeSH Major Topic]) AND Esophageal cancer[Title]) AND
(‘2000/01/01’[Date - MeSH] : ‘2020/12/01’[Date - MeSH])) AND (English[Language]) \\
            \hline
            colorectal
cancer & (((Colorectal cancer[MeSH Major Topic]) AND Colorectal cancer[Title]) AND (‘2000/01/01’[Date
- MeSH] : ‘2020/12/01’[Date - MeSH])) AND (English[Language]) \\
            \hline
            skin cancer & (((skin cancer[MeSH Major Topic]) AND skin cancer[Title]) AND (‘2000/01/01’[Date - MeSH] :
‘2020/12/01’[Date - MeSH])) AND (English[Language]) \\
            \hline
            prostate cancer & (((prostate cancer[MeSH Major Topic]) AND prostate cancer[Title]) AND (‘2000/01/01’[Date -
MeSH] : ‘2020/12/01’[Date - MeSH])) AND (English[Language]) \\
            \hline
            pancreatic
cancer & (((pancreatic cancer[MeSH Major Topic]) AND pancreatic cancer[Title]) AND (‘2000/01/01’[Date
- MeSH] : ‘2020/12/01’[Date - MeSH])) AND (English[Language]) \\
            \hline
            breast cancer & (((Breast cancer[MeSH Major Topic]) AND Breast cancer[Title]) AND (‘2000/01/01’[Date -
MeSH] : ‘2020/12/01’[Date - MeSH])) AND (English[Language]) \\
            \hline
            \end{tabular}
    \label{tab:1}
\end{table}
\subsection*{Named Entity Extraction}
Named Entity Extraction (NER), Recognizes and extracts the genes, drugs, and other related
entities from the text corpus. A user-friendly interface developed in python which browses articles
and dictionaries for entity mapping. The dictionaries prepared for each entity: diagnostics, drugs,
genes, site, risk factors, and symptoms from the portal at Cancer.Net \cite{noauthor_2008-by}, NIH
National Cancer Institute (https://www.cancer.gov/about-cancer/treatment/drugs), and Cancer
Genetics Web \cite{Pawloski2016}. Saving the extracted entities in an excel sheet
and uploaded to MySQL database for further analysis.
Mined associations from the extracted entity dataset using the Apriori algorithm \cite{Perego2001} implemented in python language.
\subsection*{Searching for Patterns
}
\begin{figure}
    \centering
    \includegraphics{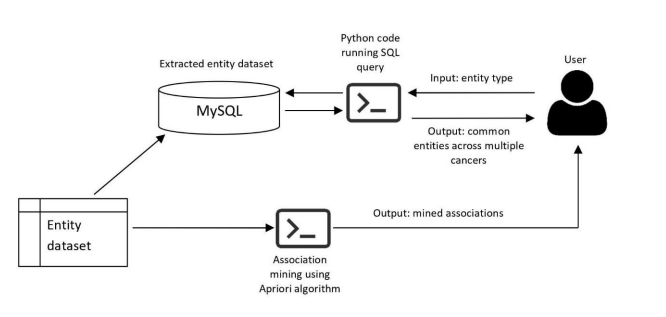}
  \caption{Association prediction workflow}
  \label{fig:fig2}
\end{figure}

The named entity extraction is quite helpful in answering some basic questions and finding the
basic knowledge in a text corpus. Further understanding necessitates studying the relationships
among entities and patterns \cite{bayardo-roberto1998}. The SQL query and apriori algorithm identify the
hidden patterns shown in Figure \ref{fig:fig2}. The apriori algorithm ranked the predicted associations based
on the Support, Confidence, and Lift score measures \cite{xie-mingyu2008}. The support measure is
how frequently an item set appears in the dataset, the confidence measure indicates how often the
rule is correct and the Lift is a ratio of the rule confidence to its expected confidence. The expected
confidence is a product of the value of the support rule body and rule head divided by the support
rule body \cite{Tan-srivastava-2002}.

\subsection{Developing Cancer Classification Model}
\begin{figure}
    \centering
    \includegraphics{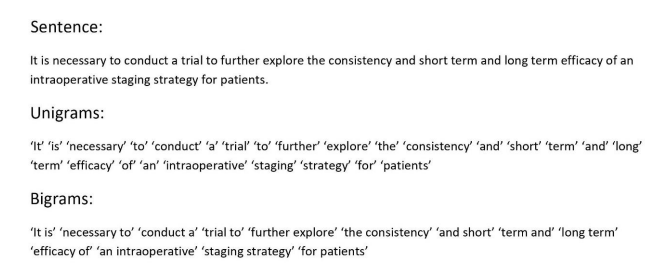}
  \caption{An example of unigrams and bigrams of a sentence.}
  \label{fig:fig3}
\end{figure}

The dataset collected via Entrez Direct software may have repeated abstracts while downloading
multiple cancers that are removed manually. For pre-processing, the NLTK python module
eliminates the unwanted elements or words from the text corpus \cite{joulin2016bag} that might hinder cancer classification. The cleaning process removes the punctuations and special characters
and stopwords/ words from abstracts that carry no meaning: words like almost, a, an, the, etc.
Performed the word-stemming of the text to treat the same word in different forms to its root form.
For example: " important, importantly, importance" are various forms of the same word
"important". Word stemming reduces the dimensions by changing the different forms to the same
base making the prediction efficient \cite{xu-jinxi-1998}. A bag of words representations \cite{Willits2014-wq}, unigrams, and bigrams as shown in Figure \ref{fig:fig3} with TermFrequency–Inverse Document Frequency (TF-IDF) scores \cite{qaiser2018} used to determine
the word importance for feature selection. In bag of words representation, the frequency of
occurrence of each word or Term-Frequency (TF) is multiplied by the Inverse Document
Frequency (IDF), and TF-IDF scores are used as feature vectors. The TF-IDF weighting score $ (W_{t,d})$ was computed by the Equation:
\begin{equation*}
    W_{t,d} = (TF_{t,d}) * log_{10}(N/DF_{t})
\end{equation*}
\newline
where $ TF_{t,d} $  refers to the frequency of the term ‘t’ occurring in an article ‘d’ \newline
$ N $  is the number of articles in the dataset
\newline
and $ DF_{t} $  refers to the number of articles containing the term ‘t’
\newline \\
The TF-IDF vectors are feature vectors for both training and test data sets. The prediction model
was built using the Linear Support Vector Classification (Linear SVC), Naïve-Bayes, Logistic
Regression, and Random Forest classifiers \cite{OsisanwoFyJET}\cite{Khreisat200972}. The SciKit Learn
python module was used to construct the model.

Naïve Bayes implemented using MultinomialNB offered by SciKit Learn and suited for text
classification. The distribution is parametrized by vectors $\theta_{y} = (\theta_{y_{1}},\theta_{y_{2}},\cdots,\theta_{y_{n}})$ for each y class, where n
is the number of features (in text classification, the size of the vocabulary) and $\theta_{y_{i}} $ is the
probability $ P(x_{i}|y) $  of feature ‘i’ appearing in a sample belonging to class ‘y’.
\newline
The parameters $ \theta_{y} $ is estimated by relative frequency counting as shown below:
\[ \theta_{y} = \frac{N_{y_{i}}+\alpha}{N_{y}+\alpha n} \]

where $ N_{y_{i}} = \sum_{x \in T} x_{i} $ i is the number of times features ‘i’ appear in a sample of y class in the training
set T and $ N_{y} = \sum_{i=1}^{n} N_{y_{i}} $ is the total count of feature class for y. The smoothing priors, $ \alpha \geq 0 $ , account
for an absence of a feature in the learning samples and prevent zero probabilities in further
computations. Linear SVC is similar to usual Support Vector Classification but supports only a
linear kernel used with the L1 regularizing parameter. Implementation of the Logistic Regression
using regularizing parameter the L2 with inverse regularization strength (C) 1.0 specifies strong
regularization and random forest classifier implemented using default parameters as given by
SciKit Learn.

\subsection{Evaluation}
Mined patterns: Predicted the entities common across the multiple cancers. The mined associations
with a minimum of three entities were identified and validated manually with online databases and
research articles for confirmation. Measured the performance of a model using the statistical
measures of accuracy, precision, and recall to validate the cancer classification model and the
patterns mined from the entities. Accuracy is the ratio of the predictions being correct, precision
is a ratio of the correct positive predictions, and recalls is a ratio of positive cases detected \cite{junker-dengel-hoch1999}. It is 'P' positive instances and 'N' negative instances in an experiment,
the four potential outcomes include True Positive (TP), True Negative (TN), False Positive (FP),
and False-Negative (FN) were calculated by:
\newline
\[Accuracy =  \frac{TP+TN}{P+N} \]
\newline
\[Precision =  \frac{TP}{TP+FP} \]
\newline
\[Recall =  \frac{TP}{TP+FN} \]
\newline
The classification model on different classifiers such as Linear SVC, Naïve Bayes, Logistic
Regression, and Random Forest) was assessed five times for the accuracy, precision, recall, and
average scores were recorded.

\subsection*{Results and Discussion
}
\begin{figure}
    \centering
    \includegraphics{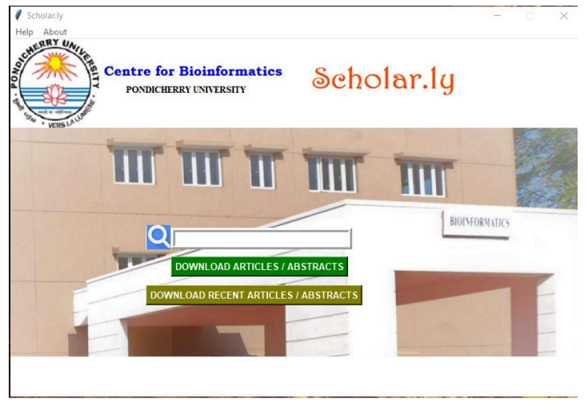}
    \caption{Homepage of Scholar.ly tool}
    \label{fig:fig4}
\end{figure}

\begin{figure}
    \centering
    \includegraphics{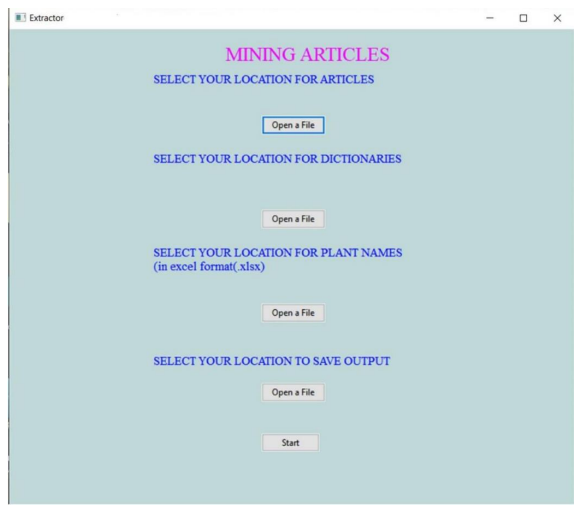}
    \caption{Homepage of Entity Extraction Interface}
    \label{fig:fig5}
\end{figure}

The study developed a mining interface for cancer entity recognition and a Machine learning
model for cancer classification.
\newline
\textbf{Mining Interface:} The scholar.ly tool shown in Figure \ref{fig:fig4} downloaded the articles and abstracts,
and the developed mining interface shown in Figure \ref{fig:fig5} extracted the cancer entities.

The entities mined from 700 cancer articles and abstracts were uploaded into the MySQL database
for pattern discovery analysis. The predicted entities common across the multiple cancers using
SQL queries and Python code are shown in Table \ref{tab:2}. The identified hidden associations and the
most significant entities with their support, confidence, and lift values are shown in Table \ref{tab:3}. The
predicted associations were validated manually with online databases such as NIH- National
Cancer Institute (https://www.cancer.gov/about-cancer/treatment/drugs), Cancer Genetics
Web \cite{Pawloski2016}, and KEGG \cite{Goto1997-pg} as well as research
articles as shown in Table \ref{tab:4}.

\begin{table}[p]
    \caption{Common entities across multiple cancers}    
    
    \begin{tabular}{p{0.35\linewidth} | p{0.1625\linewidth} | p{0.1625\linewidth} | p{0.1625\linewidth} | p{0.1625\linewidth} }
            \hline
            Common cancers & Genes & Sites & Symptoms & Risk Factors \\  
            \hline
           cervical cancer; pancreatic cancer & hepatocyte growth factor & & & \\
           \hline
           lung cancer; esophageal cancer; breast cancer & vimentin & & & \\
           \hline
           lung cancer; pancreatic cancer & palb2 & & & \\
           \hline
           gastric cancer; esophageal cancer; breast cancer & epidermal growth factor receptor & & & \\
           \hline
           esophageal cancer; colorectal cancer; pancreatic cancer & ras & & & \\
           \hline
           oral cancer; lung cancer & & lung & & \\
           \hline
           oral cancer; cervical cancer & & rectum & & \\
           \hline
            oral cancer; gastric cancer; esophageal cancer & & stomach & & \\
            \hline 
            oral cancer; pancreatic cancer; breast cancer & & breast & & \\
            \hline 
            oral cancer; gastric cancer & & liver & & \\
            \hline 
            oral cancer; cervical cancer;colorectal cancer & & colon & & \\
            \hline 
            oral cancer; lung cancer; colorectal cancer & & & discharge & \\
            \hline
           oral cancer; cervical cancer; lung cancer; colorectal cancer; skin cancer; prostate cancer; pancreatic cancer; breast cancer & & & inflammation & \\
            \hline
            cervical cancer; breast cancer & & & pain & \\
            \hline
            cervical cancer; gastric cancer & & & bleeding & \\
            \hline
            gastric cancer; prostate cancer & & & weight loss & \\
            \hline
            oral cancer; cervical cancer & & & & human papillomavirus \\
            \hline
            oral cancer; esophageal cancer & & & & tobacco \\
            \hline
            oral cancer; lung cancer; gastric cancer & & & & smoking \\
            \hline
            oral cancer; gastric cancer; esophageal cancer & & & & alcohol \\
            \hline
            cervical cancer; esophageal cancer & & & & radiation therapy \\
            \hline
            \end{tabular}
    \label{tab:2}
\end{table}

\begin{table}[p]
    \caption{Mined associations}    
    
    \begin{tabular}
{p{0.4\linewidth} | p{0.2\linewidth} | p{0.2\linewidth} | p{0.2\linewidth} | p{0.2\linewidth} }
            \hline
Associations & Support & Confidence & Lift \\
    \hline
aurora kinase a\textrightarrow biopsy\textrightarrow alcohol\textrightarrow oral
cancer & 0.153846154 & 0.857142857 & 4.178571429 \\
\hline
cisplatin\textrightarrow wnt1 inducible signaling pathway
protein 1\textrightarrow radiation therapy\textrightarrow cervical cancer & 0.08974359  & 0.5 & 5.571428571 \\
\hline
epidermal growth factor receptor\textrightarrow trastuzumab\textrightarrow gastric cancer\textrightarrow endoscopy & 0.064102564  & 0.416666667 & 6.5 \\
\hline
smoking\textrightarrow lung cancer\textrightarrow palb2 & 0.051282051 & 0.571428571 & 11.14285714 \\
\hline
ras\textrightarrow colorectal cancer\textrightarrow endoscopy & 0.064102564 & 0.833333333  & 13 \\
\hline

\end{tabular}
    \label{tab:3}
\end{table}

\begin{table}[p]
    \caption{Validated associations}    
    
    \begin{tabular}
{p{0.5\linewidth} | p{0.5\linewidth} }
            \hline
Associations & Validation \\
    \hline
aurora kinase a\textrightarrow biopsy\textrightarrow alcohol\textrightarrow oral
cancer & PMID: 25697104, American
cancer society \\
\hline
cisplatin\textrightarrow wnt1 inducible signaling pathway
protein 1\textrightarrow radiation therapy\textrightarrow cervical cancer & cancer-genetics, KEGG,
American cancer society,
PMID: 30651114
 \\
\hline
epidermal growth factor receptor\textrightarrow trastuzumab\textrightarrow gastric cancer\textrightarrow endoscopy &  PMID: 33990570, cancergenetics, American cancer society,
KEGG\\
\hline
smoking\textrightarrow lung cancer\textrightarrow palb2 & PMID: 29387807 \\
\hline
ras\textrightarrow colorectal cancer\textrightarrow endoscopy &  PMID: 29534749, cancergenetics \\
\hline

\end{tabular}
    \label{tab:4}
\end{table}

\begin{figure}
    \centering
    \includegraphics{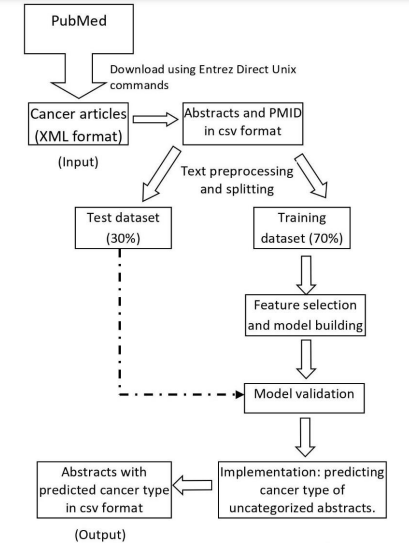}
    \caption{Cancer classification workflow}
    \label{fig:fig6}
\end{figure}

\textbf{Classification Model:}  The abstracts of ten different cancer types (32, 200) were downloaded from
PubMed by the search query and pre-processed for training data preparation. Four classifiers
(Linear SVC, Naïve Bayes (NB), Logistic Regression (LR), and Random Forest (RF)) evaluated
the accuracy, recall, and precision for the test dataset of 13,800 abstracts. Figure \ref{fig:fig6} is the workflow
of the cancer classification model and recorded the performance scores in Table \ref{tab:5}. The Linear
SVC classifier provided the best performance scores and made very accurate predictions when
implemented on 100 abstracts across multiple cancer types. The predicted cancer labels by the
model compared with the gold-standard cancer labels obtained from MeSH (Medical Subject
Headings) provided a good correspondence with the actual cancer labels. It concludes that the
classification model predicted the cancer types appropriately.

\begin{table}[p]
    \caption{Classifier performance scores}    
    \begin{threeparttable}
    \begin{tabular}
{p{0.2\linewidth} | p{0.2\linewidth} | p{0.2\linewidth} | p{0.2\linewidth} | p{0.2\linewidth} }
            \hline
SCORES (Avg.) & Linear SVC\tnote{1} & NB\tnote{2} & RF\tnote{3} & LR\tnote{4} \\
\hline
ACCURACY & 0.991 & 0.927 & 0.982 & 0.984 \\
\hline
RECALL & 0.990 & 0.908 & 0.970 & 0.986 \\
\hline
PRECISION & 0.987 & 0.941 & 0.985 & 0.988 \\
\hline
TIME TAKEN (in seconds) & 71 & 72 & 134 & 91 \\
\hline
\end{tabular}
    \label{tab:5}
\begin{tablenotes}
\item [1] \textbf{Linear Support Vector Classifier} 
\item [2] \textbf{Naïve Bayes} 
\item [3] \textbf{Random Forest}
\item [4] \textbf{Logistic Regression} 
\end{tablenotes}
\end{threeparttable}
\end{table}

\section{Future Work}
In addition to the association prediction, network construction for diseases of different cancer types
is possible using the predicted patterns. It gives a deep insight into the various factors responsible
for cancer progression. The network analyzes the gene-protein interaction and drug-protein
interactions that may be useful to uncover the potentially novel associations that are yet to be
studied. 

\section{Conclusion}
The computational mining interface for named entity recognition and cancer classification model
has been developed. Out of examination, the Support Vector Machine outperformed other
classifiers. The 'scholar.ly' tool reduces the researcher's effort to a great extent to collect the articles
in a single location from PubMed. The automated mining interface was developed generically,
user-friendly dictionary-based, and tested on extracting the cancer domain. The extracted entities
were analyzed to predict the hidden patterns that might lead to novel associations among entities.
Scalability is the main feature of the interface as there was no limit on the number of articles and
the number of extracted entities. The study concludes that the interface contributes significant
benefits in the research field of medical, chemical, and various domains to predict novel scenarios. 

\subsection*{Recognition}
This paper got accepted for paper presentation at the International Conference on Knowledge
Discoveries on Statistical Innovations and Recent Advances in Optimization (ICON-KSRAO) on
29th and 30th December 2022. 

\printbibliography

\end{document}